\documentclass{uai2022} % for initial submission
\usepackage[american]{babel}
\usepackage{natbib} % has a nice set of citation styles and commands
\usepackage{mathtools} % amsmath with fixes and additions
\usepackage{booktabs} % commands to create good-looking tables

\usepackage{amsfonts}

\usepackage{tikz} % nice language for creating drawings and diagrams
\usetikzlibrary{arrows,shapes,plotmarks,decorations.pathmorphing,automata}
\usetikzlibrary{backgrounds,calc,positioning,fadings}

\tikzset{every picture/.style={node distance=2cm,shorten >= 1pt, shorten <= 1pt}}
\tikzset{>=stealth'} 
\tikzstyle{graphnode} = 
   [circle,draw=black,minimum size=22pt,text centered,text
     width=22pt,inner sep=0pt] 
\tikzstyle{var}   =[graphnode,fill=white]
\tikzstyle{obs}   =[graphnode,fill=black,text=white]
\tikzstyle{act}   =[rectangle,draw=black,text=white,minimum
size=22pt,text centered, text width=22pt,inner sep=0pt]
\tikzstyle{fac}   =[rectangle,draw=black,fill=TUgold,minimum size=5pt]
\tikzstyle{facprior} =[rectangle,draw=black,fill=black,text=white,minimum size=5pt]
\tikzstyle{edge}  =[draw=white,double=black,thick,-]
\tikzstyle{prior} =[rectangle, draw=black, fill=black, minimum size=
5pt, inner sep=0pt]
\tikzstyle{dirprior} = [circle, draw=black, fill=black, minimum
size=5pt, inner sep=0pt]

\newcommand{\emilia}[1]{{\bf \color{blue!50!black} [EMILIA: #1]}}
\newcommand{\Rbb}{\mathbb{R}}
\newcommand{\Nbb}{\mathbb{N}}
\newcommand{\Xbb}{\mathbb{X}}

\newcommand{\Gcal}{\mathcal{G}}

\newcommand{\Hcal}{\mathcal{H}}

\newcommand{\Lcal}{\mathcal{L}}

\newcommand{\diff}{\,\text{d}}
\usepackage[noabbrev, capitalize]{cleveref}
\usepackage{comment}
\usepackage{caption}
\usepackage{subcaption}

\DeclareMathOperator{\NO}{NO}
\DeclareMathOperator{\Id}{Id}

\title{Approximate Bayesian Neural Operators: Uncertainty Quantification for Parametric PDEs}

\author[1]{\href{mailto:<emilia.magnani@uni-tuebingen.de>}{Emilia~Magnani}{}}
\author[1]{Nicholas~Kr\"amer}
\author[1]{Runa~Eschenhagen}
\author[3,4]{Lorenzo~Rosasco}
\author[1,2]{Philipp~Hennig}
\affil[1]{%
    University of T\"ubingen\\
    Germany
}
\affil[2]{%
    Max-Planck-Institute for Intelligent Systems\\
    T\"ubingen\\
    Germany
}
\affil[3]{%
    MaLGa Center - DIBRIS \\
    University of Genova and Istituto Italiano di Tecnologia\\
    Genova \\
    Italy
  }
 \affil[4]{%
   CBMM\\
   Massachusetts Institute of Technology\\ 
   Cambridge\\
   MA, USA
  }
  
\begin{document}
\maketitle

\begin{abstract}
 Neural operators are a type of deep architecture that learns to solve (i.e. learns the nonlinear solution operator of) partial differential equations (PDEs). The current state of the art for these models does not provide explicit uncertainty quantification. This is arguably even more of a problem for this kind of tasks than elsewhere in machine learning, because the dynamical systems typically described by PDEs often exhibit subtle, multiscale structure that makes errors hard to spot by humans. In this work, we first provide a mathematically detailed Bayesian formulation of the ``shallow'' (linear) version of neural operators in the formalism of Gaussian processes. We then extend this analytic treatment to general deep neural operators using approximate methods from Bayesian deep learning. We extend previous results on neural operators by providing them with uncertainty quantification. As a result, our approach is able to identify cases, and provide structured uncertainty estimates, where the neural operator fails to predict well.
\end{abstract}

\section{Introduction}
Neural operators  \citep{li2020neural,li2020fourier,li2020multipole,li2021markov,kovachki2021neural} are a deep learning architecture tailored to reconstruction problems related to partial differential equations.
They approximate mappings between infinite-dimensional vector spaces of functions, such that -- once trained -- solutions of entire families of parametric partial differential equations (PDEs) can be represented by a single neural network.
% They are very powerful because they can approximate mappings between infinite dimensional spaces (operators) in a fast way. \nico{}
% Moreover,  once trained, they can learn a whole family of PDEs without the necessity of re-training the network parameters.
This process is subject to several sources of uncertainty, which can result in a potentially significant prediction error because of the nonlinear -- and nonintuitive -- interactions of different stages of the approximation. The goal of this paper is to develop methods for estimating this error at a practically acceptable computational cost.
This kind of functionality is urgently needed in this domain: Due to the intricate and often not intuitive nature of the dynamical systems described by PDEs, it can be hard for the human eye to detect prediction errors, even when they are large. 

In this paper, we develop an approximate Bayesian framework for neural operators -- from a theoretical, and a computational point of view. We begin with a brief review of neural operators.
Then, using linear, parametric PDEs as guiding examples, we show how their ``shallow'' (single-layer) base case allows an analytic Bayesian treatment in the formalism of Gaussian processes (\cite{GpRasmWill}). 
Even though PDEs are linear, the parameter-to-solution operators are not.
Although this linear case is primarily of theoretical interest, it forms a core contribution of this paper that may make this model class more easily accessible to the Bayesian machine learning community. We then extend the theoretical analysis to the deep case. Here, analytic treatments are no longer possible, so we fall back on approximations developed for Bayesian deep learning. Specifically, we focus on Laplace approximations \citep{mackay1992evidence} which are easy to add post-hoc even to pretrained networks, and add only moderate computational cost relative to deep training without uncertainty quantification \citep{LaplaceRedux}. We show in experiments that the resulting method can capture structure in predictive error both in the over- and under-sampled regime.
In \cref{method} we discuss some theoretical background and develop a probabilistic framework for neural operators. 
We discuss the related work in \cref{sec:related_work}.
In \cref{sec:experiments} we provide empirical results.
\begin{figure*}[ht] 
      \begin{center}
            %\framebox[4.0in]{$\;$}
            \includegraphics[width=1\textwidth]{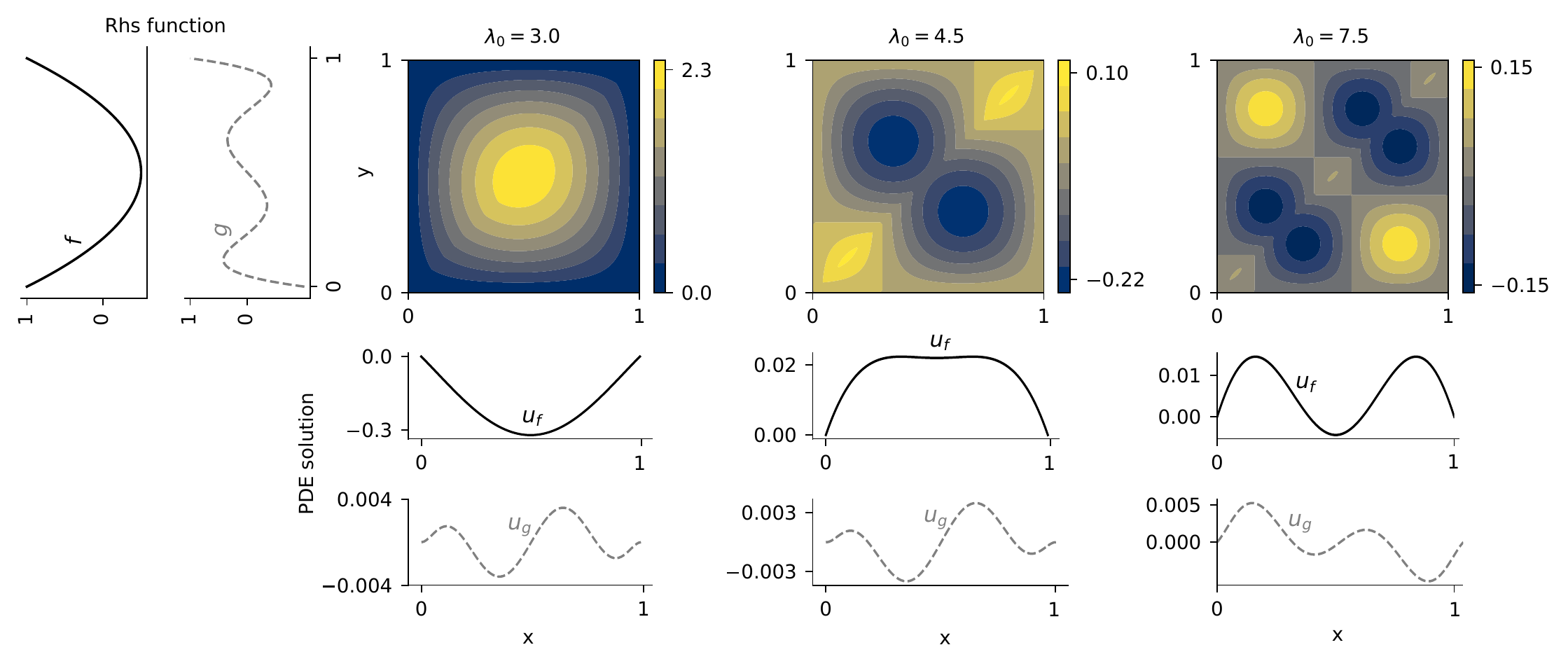}
      \end{center}
      \caption{Green's functions in \cref{eq:greens_function} for different values of $\lambda_0 = \{3, 4.5, 7.5 \}$.  On the left, 
      right-hand-side functions $f$, $g$ for the PDE in \cref{eq:test_problem} and respective solutions 
      $u_f$, $u_g$ for the correspondent $\lambda_0$-value, computed through \cref{eq:fund_sol}.
      %Note how $G_{\lambda_0}$ is not a continuous function of $\lambda_0$.
      }\label{fig:greens_function_and_solution}
\end{figure*}

\section{METHOD }\label{method}

\subsection{PDEs And Green's Function} \label{pdes_green}
One of the main fields of applications of neural operators are partial differential equations (PDEs). 
In this work we consider the family of parametric PDEs
\begin{equation}
\label{eq:general_test_problem}
\begin{split}
\big(\Lcal_\lambda u \big) (x) 
& = f(x), \ \ \  x \in D \\
u(x) &= 0, \ \ \ \ \ \ \ \  x \in \partial D 
\end{split}
\end{equation}
for some sufficiently well-behaved, bounded domain $D \subset \mathbb{R}^d$ with boundary $\partial D$ (e.g.\ open, bounded $D$ with Lipschitz boundary $\partial D$), where $U \ni u \colon D \rightarrow \Rbb$,  $F \ni f \colon D \rightarrow \Rbb$, $\lambda \in \Lambda$, with $U$, $F$ and $\Lambda$ appropriate function spaces.
The precise nature of those function spaces is not important for the remainder of this work.
The function $\lambda$ parametrises the differential operator $\mathcal{L}_\lambda$.

\cref{eq:general_test_problem} defines a solution operator
\begin{align} \label{eq:neurop_generalized}
\mathcal{H} \colon \Lambda \times F  \rightarrow U, \quad (\lambda,f) \mapsto u_{\lambda,f}
\end{align}
in the sense that  $\Hcal(\lambda,f)(x) = u_{\lambda,f}(x)$ solves the PDE  for the given functions $\lambda$ and $f$. 
Even though the PDE is linear, $\Hcal$ is (possibly highly) nonlinear.
In particular, here we consider the case where $\lambda$ is fixed, so the solution operator can be written as
\begin{equation}\label{eq:G-operator}
   \mathcal{G} \colon f \mapsto u . 
\end{equation}
The operator $\Gcal$, like $\Hcal$, is a map between function spaces. The idea behind neural operators is to approximate the operator $\Gcal$ (or $\Hcal$)  with a single neural network trained on function observations ${\{f_i,  u_i \}}_{i=1}^N$.
 Thus, instead of approximating the solution of the PDE for only a fixed $f$, neural operators directly infer the operator $\Gcal$.
 
 Numerically, the functions $f$ and $ u$ are observed on a discretisation grid of the function domains.
 Considering the operator in \cref{eq:G-operator} is a key step to understand the learning process of neural operators. In fact, observe how  $\mathcal{G}$ is the inverse of the operator $\mathcal{L}_\lambda$.
 The neural operator is therefore learning an operator, $\Gcal$, through function observations
 ${\{f_i,  u_i \}}_{i=1}^N$ that derive from the action of its inverse. In other words, during training, the neural operator is implicitly learning to invert the differential operator $\Lcal_\lambda$.
In particular, in the case where the differential operator is linear and admits a Green's function $G$,  the solution of \cref{eq:general_test_problem} can be expressed through integration with the kernel $G$ 
 \begin{equation}\label{eq:fund_sol}
 u_{\lambda}(x) = \int_D G_{\lambda}(x,y) f(y) \diff y.
 \end{equation}
Hence, learning the operator $\Gcal$ is here equivalent to learn  the function $G$ which means that an operator-learning task can be reduced to that of function-reconstruction.
The structure of neural operators in its one-layer case is inspired by the Green's solution formula for linear PDEs in \cref{eq:fund_sol}. We will examine their architecture, in the more general case, in the next section.

In the general analysis of linear PDEs
(we refer to e.g.\ \citet{evans10}
for background on PDEs), the Green’s function $G(x,y)$ represents the impulse response of the linear operator $\mathcal{L}_\lambda$, that is $\mathcal{L}_\lambda (G)(\cdot,y)= \delta(\cdot-y)$ for $y \in D$, where $\delta$ denotes the Dirac delta distribution.
 Note how  $\mathcal{L}_\lambda$ is a linear operator,
 whereas the Green's function is usually nonlinear in either arguments.
To visualize the presented concepts, we consider  the boundary value problem 
\begin{equation}\label{eq:test_problem}
\begin{split}
\big(-\Delta - \lambda_0^2 \Id \big)\
  u (x) &= f(x), \ \ \  x \in [0,1],  \\
u(0)=u(1) &= 0,
\end{split}
\end{equation}
that
admits a Green's function in closed form,
\begin{equation}      \label{eq:greens_function}  
      G_{\lambda_0} (x, y)  :=                   
            \frac{A + B}{\lambda_0 \sin (\lambda_0)} 
\end{equation}
where we abbreviated
\begin{align}
A &:= H(y-x)\sin (\lambda_0 x)
                  \sin (\lambda_0(1-y))\\ 
B &:= H(x-y) \sin (\lambda_0(1-x)) \sin (\lambda_0 y),
\end{align}
and $H$ denotes the Heaviside step function.
\cref{eq:test_problem} relates to \cref{eq:general_test_problem} in the sense that the differential operator $\Lcal_{\lambda_0} = 
(-\Delta - \lambda_0^2 \Id)$ is parametrised by $\lambda_0$.
Green's functions $G_{\lambda_0}$ for different values of $\lambda_0$, as well as the solutions computed through the solution formula in \cref{eq:fund_sol}, are depicted in Figure
 \ref{fig:greens_function_and_solution}.
 
 %$\mathcal{F} \colon f \mapsto u $  that map $f$, instead of $\lambda$, to $u$.
%In our evaluation, we set $\lambda(x) \equiv \lambda_0^2$ for some $\lambda_0 \in \Rbb$, but define the concepts in the general case, in order to establish mathematical rigour with respect to the domains of the operators.

 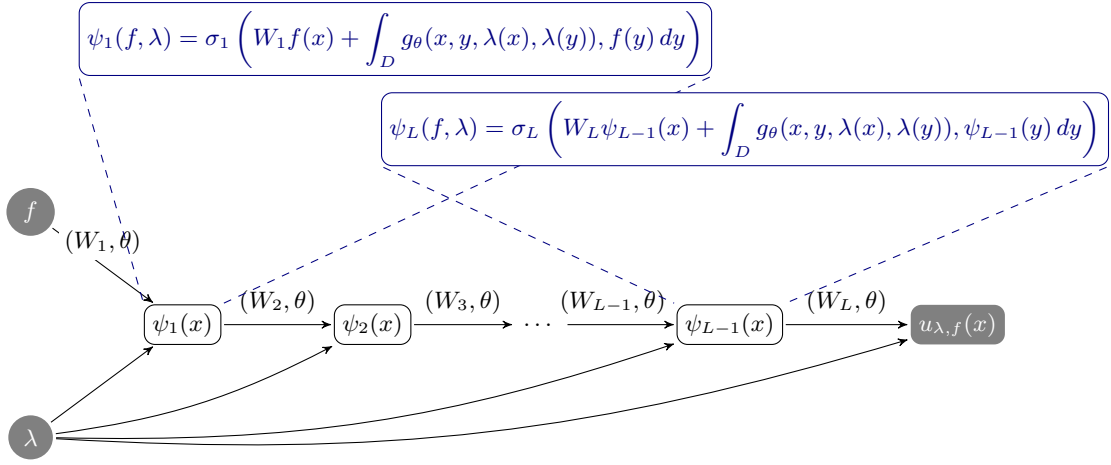
\begin{figure*}[!ht] 
      \begin{center}\small
            %\framebox[4.0in]{$\;$}
            % \includegraphics[width=.9\textwidth]{fig:neurop_architecture.jpeg}
            %!TEX root=./iclr2022_conference.tex
\begin{tikzpicture}
	\node[shape=circle,fill=gray,text=white] (f) at (-1,-.5) {$f$};
	\node[shape=circle,fill=gray,text=white] (l) at (-1,-3.5) {$\lambda$};

	\node[shape=rectangle,rounded corners,text=black,draw=black] (psi1) at (1,-2) {$\psi_1(x)$};
	\node[shape=rectangle,rounded corners,text=black,draw=black] (psi2) at (3.5,-2) {$\psi_2(x)$};
	\node (psi3) at (5.7,-2) {$\dots$};
	\node[shape=rectangle,rounded corners,text=black,draw=black] (psiL) at (8.2,-2) {$\psi_{L-1}(x)$};
	\node[shape=rectangle,rounded corners,fill=gray,text=white] (u) at (11.2,-2) {$u_{\lambda,f}(x)$};

	\path[->] (f) edge node[above,fill=white,inner sep=2pt,yshift=2pt] {$(W_1,\theta)$} (psi1)
		(l) edge node[above,fill=white,inner sep=2pt,yshift=2pt] {} (psi1)
		(psi1) edge node[above,fill=white,inner sep=2pt,yshift=2pt] {$(W_2,\theta)$} (psi2)
		(psi2) edge node[above,fill=white,inner sep=2pt,yshift=2pt] {$(W_3,\theta)$} (psi3)
		(psi3) edge node[above,fill=white,inner sep=2pt,yshift=2pt,xshift=-3pt] {$(W_{L-1},\theta)$} (psiL)
		(psiL) edge node[above,fill=white,inner sep=2pt,yshift=2pt] {$(W_L,\theta)$} (u)
		(l) edge[bend right=10] (psi2)
		(l) edge[bend right=10] (psiL)
		(l) edge[bend right=10] (u);

	\node[shape=rectangle, rounded corners, text=blue!50!black,draw=blue!50!black] (l1) at (3.8,1.8) {$\displaystyle\psi_1(f,\lambda) = \sigma_1\left(W_1f(x) + \int_D g_\theta(x,y,\lambda(x),\lambda(y)),f(y)\,dy\right)$};

	\draw (l1.south east) edge[dashed,draw=blue!50!black] (psi1.north east);
	\draw (l1.south west) edge[dashed,draw=blue!50!black] (psi1.north west);

	\node[shape=rectangle, rounded corners, text=blue!50!black,draw=blue!50!black,fill=white] (l2) at (8.4,0.6) {$\displaystyle \psi_L(f,\lambda) = \sigma_L\left(W_L \psi_{L-1}(x) + \int_D g_\theta(x,y,\lambda(x),\lambda(y)),\psi_{L-1}(y)\,dy\right)$};

	\draw (l2.south east) edge[dashed,draw=blue!50!black] (psiL.north east);
	\draw (l2.south west) edge[dashed,draw=blue!50!black] (psiL.north west);

\end{tikzpicture}
      \end{center}
      \caption{ Neural operator architecture $\NO_\Theta$. Each layer $l$ computes a new function $\psi_l$, that contains the neural network $g_\theta$ in the integrand. Parameters of the layers are depicted on the respective arrows. The input function $f$ enters as an ''initialisation'' only in the first layer, whereas the function $\lambda$ enters in $g_\theta$ at every $\psi_l$.  }\label{fig:noparchitecture}
\end{figure*}
 %
  %Green's functions are a powerful tool for solving nonhomogeneous linear PDEs because they express the solution of a boundary value problem explicitly in terms of the Green's function. The most common linear operators have a known fundamental solution, but in general it is  very challenging to such solutions for non-linear operators. 

\subsection{Neural Operator Essentials}
\label{sec:neural_operators}
Before formulating a Bayesian framework for neural operators, we recall their structure.
A more thorough explanation of what follows can be found in the work by \citet{li2020neural,li2020fourier,li2020multipole,li2021markov,kovachki2021neural}. 

A neural operator is a neural network architecture designed to approximate the general
solution operator $\mathcal{H}$ in
 \cref{eq:neurop_generalized}. 
% We next describe the concepts in the most general case of the operator $\mathcal{H} \colon (\lambda,f) \mapsto u$.
 For particular cases, such as the operator $\Gcal$ in \cref{eq:G-operator} where $\lambda$ is fixed, or for operators mapping $\lambda \mapsto u$ (where $f$ is fixed), an analogous construction is straight forward.

Let $g_\theta: D \times D \times \Rbb \times \Rbb \rightarrow \Rbb$ be a neural network with parameters $\theta$.  %Any choice is possible.
Define the neural operator $\NO_\Theta$ as a composition of $L \in \Nbb$ layers
\begin{align}\label{neur_op}
      \NO_\Theta \colon \Lambda \times F  &\rightarrow U,\notag \\
      (\lambda,f) &\mapsto (\psi_L \circ \psi_{L-1} \circ ...\circ \psi_1)(\lambda, f),
\end{align}
where each layer 
\begin{align}
    \psi_\ell: \Phi \rightarrow \Phi, \quad \ell = \{ 1, \dots L\},
\end{align}
is defined as a composition of (i) integrating the output of the previous layer against $g_\theta$, and (ii) combining the integral with a linear component and an activation function $\sigma$,
\begin{multline}\label{neurop_iteration}
        \psi_\ell(g)(x) =\\ 
       \sigma \bigg( W_\ell g(x)  
       + \int_D g_\theta(x, y, \lambda(x), \lambda(y) )  g(y) \diff y\bigg).
\end{multline}
$\Phi$ is a vector space of functions mapping from $D$ to $\Rbb$. The final layer of the neural operator maps into $ U$, $\psi_L \colon \Phi \rightarrow U$.
In practice, the integral cannot be computed in closed-form and a suitable quadrature formula needs to be employed (which turns the integral into a  weighted sum of evaluations of the integrand; see e.g.\ \citep{davis2007methods}).
The parameters $\Theta$ of $\NO_\Theta$ include the parameters $\theta$ of $g_\theta$ as well as the weights in 
each layer $W_\ell$, i.e. $\Theta = \theta \cup {\{ W_\ell \}}_{\ell=1}^{L}$.
Loosely speaking, one can think of this construction as a deep neural network ($\NO_\Theta$) that iteratively approximates the solution $u_{\lambda, f}$ (see \cref{eq:neurop_generalized}) and at every iteration (layer) employs another neural network ($g_\theta$). For a visualisation of $\NO_\Theta$ see Figure \ref{fig:noparchitecture}.

This architecture is inspired by the process of solving linear PDEs with Green's functions:
In the case where $L = 1$, $\lambda \equiv \lambda_0^2$, $\sigma = \Id$, and $W_1 = 0$, and we consider the mapping  $\mathcal{G} \colon f \mapsto u $, the neural operator 
approximating $\mathcal{G}$ becomes
\begin{align}\label{eq:one_layer_neurop}
      \NO_\Theta(f) = \NO_\theta(f) = \int_D g_\theta(x, y) f(y) \diff y.
\end{align}
If $g_\theta$ is a sufficiently accurate approximation of the Green's function $G_{\lambda_0}$ in \cref{eq:greens_function}, \cref{eq:one_layer_neurop} is the solution formula of the PDE in \cref{eq:test_problem}. In the next section we will provide a probabilistic formulation of the one layer architecture in \cref{eq:one_layer_neurop} that is based on the formalism of Gaussian processes.
% An alternative architecture based on solving linear PDEs with a Fourier transform has been proposed by \cite{li2020fourier}.

Note how $\NO_\Theta$ approximates an operator. While, technically speaking, this means that its training and test set consist of  functions, in the numerical computation, these functions need to be observed on some grid. 
Let $\{\lambda_1, ..., \lambda_N \} \times \{f_1, ..., f_M\}$ be a set of training inputs, each of which shall be observed on some mesh $\Xbb := \{x_1, ..., x_K\}$.
In total, that makes $NK\times MK=NM K^2$ training inputs.
Without loss of generality, and for the sake of simple notation, assume that the solution  of the PDE and the respective inputs are observed on the same mesh $\Xbb$.
Thus, we observe $NM$ solutions $u_{11}, ..., u_{NM}$, i.e. $NMK$ training outputs -- one set of evaluations at $\Xbb$ for each solution $u_{nm}$ associated with $(\lambda_n, f_m)$, $n=1, ..., N$, $m=1, ..., M$.
Each of these outputs is a function that maps from $D$ to $\Rbb$, thus $u_{nm}(\Xbb) \in \Rbb^{K}$.
The relation between inputs and outputs is
\begin{equation}
      u_{nm} = \Hcal(\lambda_n, f_m) {\approx}\NO_\Theta(\lambda_n, f_m).
\end{equation}
While this equation is between functions, once discretised, it becomes an equation between vectors.
%because $\lambda_n$, $f_m$ and $u_{nm}$ are functions.
To be able to optimise the parameters, we introduce the loss function
\begin{align}
      \mathcal{L}: \Rbb^{K} \times \Rbb^{K} \rightarrow [0, \infty).
\end{align}

The network parameters $\Theta$ are then computed by (approximately) solving the minimisation problem
\begin{align}
      \Theta^* = \arg \min_\Theta \sum_{n,m}\mathcal{L} (u_{nm}(\Xbb), \NO_\Theta(\lambda_n, f_m)(\Xbb)),
\end{align}
where we used the above vectorised notation.
This minimisation can be carried out with any of the optimisers popular in deep learning (see e.g. \citep{le2011optimization}).
Note that by approximating directly the solution operator $\Hcal$, $\NO_\Theta$ simultaneously learns the entire family of PDEs parametrised by $f, \lambda$ without the need
 of re-training the network for a new $\lambda$ or $f$.
Considering that these new inputs samples can be out of distribution cases, which are notoriously harder to predict \citep{hendrycks17baseline}, it is even more important to introduce uncertainty quantification for these architectures.

\subsection{Bayesian Neural Operators and Gaussian processes} \label{subsec:gp_regression}
Here we develop the Bayesian probabilistic framework for the neural operator. We start by observing that the special case of a one-layer network allows an analytic and in fact non-parametric Bayesian treatment through a Gaussian process model. This setting provides not just a useable algorithm, but also an important conceptual base-case that is not prominently discussed in previous works on neural operators (including non-Bayesian ones). In \cref{subsec:llla_neurop}, this “shallow” treatment will be extended to the deep setting using a linearisation in form of the Laplace approximation, which again provides a Gaussian posterior distribution, albeit an approximate one.
%We start with analyzing the concepts in the  one-layer network case of \cref{eq:one_layer_neurop}, and then extend it to the $L$-layers architecture of \cref{neur_op} in the next section. A natural and powerful way to formulate the Bayesian treatment in the one-layer setting is through the non-parametric framework of Gaussian process (GP) regression. While this approach only works in this ``shallow'' setting, we believe it is valuable for providing an intuition, especially for readers versed in Bayesian machine learning. In the next section, we corroborate our analytical construction with experiments (for both shallow and deep setting) that explicitly provide uncertainty estimates and posterior means of PDE solutions allowing to detect when the neural operator is wrong in its prediction.

Consider the solution operator $\mathcal{G} \colon f \mapsto u $ of the \emph{linear} PDE in
\cref{eq:test_problem}.  
In this case  $\mathcal{G}$ can be approximated with a one-layer neural operator, that in its single iteration computes the PDE solution as the integral 
 \begin{equation}\label{eq:onelayer_neurop_again}
  \NO_\theta = u_f(x) = \int_D g_\theta(x, y) f(y) \diff y.
  \end{equation}
  As observed in Section \ref{pdes_green}, this ``shallow'' form of the neural operator is based on Green's solution formulas
  for linear PDEs. 
  %see \cref{eq:fund_sol}.
Since the considered linear PDE admits an analytic Green's function $G$ (see \cref{eq:greens_function}),
and since the only parameters of $\NO_\Theta$ are the ones of the neural network $g_\theta $, i.e. $\Theta = \theta$,  learning the operator $\mathcal{G}$ is here equivalent to learning the function $G$.
Therefore, for this setting, one can reformulate the task of inferring the solution operator $\mathcal{G} \colon f \mapsto u $ (which maps between infinite-dimensional vector spaces of functions) as the
inference problem of learning the function $G \colon \Rbb^2 \rightarrow \Rbb$. 

In contrast to conventional GP regression, instead of direct observations of $G$, we only have access to $G$ through the integrals
  $u_n = \int_D G(x, y) f_n(y) \diff y $ for every data point $f_n$, $n= 1, \dots, N$. 
  We define the  integral
  operator $\mathcal{A}_f = \mathcal{A}$  acting on $G$ as $\mathcal{A} G = \int_D G(\cdot, y) f(y) \diff y = u (\cdot) $.
  Since $\mathcal{A}$ is a linear operator, a Gaussian likelihood involving these observations (including the limit case of noise-free observations) remains conjugate to a GP prior and a Gaussian posterior can be computed in closed-form \citep{tanskanen2020non,longi2020sensor}. 
  
  Assume a Gaussian prior $G \sim \mathcal{GP}(\mu, k_{\theta}) $ with mean function $\mu \colon \mathbb{R}^2 \rightarrow \mathbb{R}$ and a parametrised kernel function $k_{\theta} \colon \mathbb{R}^2 \times \mathbb{R}^2 \rightarrow \mathbb{R} $. Assuming 
$u \mid G \sim \mathcal{N} (\mathcal{A} G, {\sigma^2})$, the posterior distribution over $G$ is a Gaussian process with mean and covariance
\begin{equation} \label{eq:gp_post}
      \begin{split}
            & \mathbb{E}[G] = \mu + \mathcal{A}^\ast k_{\theta}{(\mathcal{A}\mathcal{A}^\ast k_{\theta} + \sigma^2)}^{-1} (u -\mathcal{A} \mu) \\
            &\text{Cov}(G) = k_{\theta} - \mathcal{A}^\ast k_{\theta}{(\mathcal{A}\mathcal{A}^\ast k_{\theta} + \sigma^2)}^{-1} \mathcal{A} k_{\theta}.
      \end{split}
\end{equation}
where $\mathcal{A}^\ast$ is the adjoint of $\mathcal{A}$.
With the posterior distribution over $G$ at hand we can compute uncertainty estimates on the prediction, draw posterior samples, and exploit all the other properties of GP regression. 
Moreover, the versatility of GPs
 allows to include prior information about $G$ in the kernel $k_\theta$. For example, the fact that Green's functions are symmetric, i.e. $G(x,y) = G(y,x)$, can be encoded in  $k_\theta$ (\cite{duvenaud2014automatic}).
Since the solution $u$ is a linear function of $G$,
the Gaussian posterior over $G$ induces
a GP over the solution $u$. That is, we obtain a probabilistic estimate over the PDE solution. Moreover, since we learned the solution operator  $\mathcal{G} \colon f \mapsto u $, we directly obtain an estimate of all the PDE solutions for new right hand side functions $f^\ast$.
In \cref{subsec:gp_experiments} we use this GP regression framework to learn the solution operator of \cref{eq:test_problem}.

\subsection{From GP To NN: Last-Layer Laplace Approximation On Neural Operators}
\label{subsec:llla_neurop}
%\emilia{Write here something along the lines of: 
%In the previous section we built an analytic Bayesian %framework for the shallow neural operator. We will show in %experiments how this is useful. Now, for the deep case, we %can exploit approximate methods from Bayesian deep %learning. In particular, we choose a linearisation with %Laplace approximations, because they are powerful and %cheap, and because they provide a "principled way to %translate exact shallow models to approximate deep ones" %(->how? ref?)}

While we can directly use GP regression to obtain uncertainty estimates on PDE solutions for the one-layer neural operator, this approach cannot be directly applied to deep neural operators, which contain non-linearities. However, we can use approximate inference techniques from Bayesian deep learning to obtain an approximation to the posterior distribution
over the weights $p(\Theta \! \mid \! \mathcal{D})$ with $\mathcal{D} = \{\lambda_n, f_m, u_{nm}\}$, for $n=1, \dots , N$ and $m=1, \dots , M$.
%The parameters $\Theta,$ include the weights $\theta$ of $g_\theta$ and a set of weights for every layer of the network $ {\{ W_\ell \}}_{\ell=1}^{L}$.
Since the computation of the true posterior is intractable, it is common to use a Gaussian approximation \citep{mackay1992evidence, blundell2015weight}. 
To make predictions with the approximate posterior $q(\Theta),$ we need the predictive distribution
\begin{equation}
\label{eq:predictive}
\begin{split}
    p(u_\ast &\mid \NO_\Theta (\lambda_\ast, f_\ast), \mathcal{D} ) \approx \\ &\int p(u_\ast \mid \NO_\Theta (\lambda_\ast, f_\ast)) q(\Theta)  \diff \Theta
\end{split}
\end{equation}
for test functions $(\lambda_\ast, f_\ast)$.
In general, computing this predictive distribution requires further approximation, such as the local linearisation of the neural network \citep{immer2020improving} which results in a Gaussian predictive distribution for a Gaussian likelihood.
Alternatively, we can use a Laplace approximation, a relatively simple and early form of Bayesian deep learning  \citep{mackay1992evidence}, on only the last layer of the network.
This allows us to apply Laplace approximations to the intricate architecture of neural operators for efficient uncertainty quantification.

The Laplace approximation for neural networks requires a maximum a-posteriori (MAP) estimate which is obtained by
minimizing the loss $\Lcal (\mathcal{D};\Theta)$
\begin{equation}
\begin{split}
    \Theta_{\text{MAP}} &= \arg\min_{\Theta} \Lcal (\mathcal{D};\Theta) \\
    &= \arg\min_{\Theta} \big( r(\Theta) + \sum_{n, m} \ell(\lambda_n, f_m, u_{nm}, \Theta) \big).
\end{split}
\end{equation}

The empirical risk $\ell(\lambda_n, f_m, u_{nm}, \Theta)$ corresponds to the negative log likelihood $-\log p(u_{nm} \mid \NO_\Theta(\lambda_n, f_m))$ and the regularizer $r(\Theta)$ to the negative log prior distribution $-\log p(\Theta).$
The general idea of the Laplace approximation is to construct a local Gaussian approximation to the posterior $p(\Theta \mid \mathcal{D})$ by using a second order expansion of the loss $\mathcal{L}(\mathcal{D};\Theta)$ around $\Theta_{\text{MAP}}$
\begin{equation}
\begin{split}
   & \Lcal (\mathcal{D};\Theta)  \approx 
   \Lcal(\mathcal{D};\Theta_{\text{MAP}})
 \\  
   & 
    + \frac{1}{2}{(\Theta - \Theta_{\text{MAP}} )}^\text{T} \big({\nabla_\Theta^2 \mathcal{L} (\mathcal{D};\Theta)|}_{\Theta_{\text{MAP}}} \big)
    (\Theta - \Theta_{\text{MAP}} ),
\end{split}
\end{equation}
where the first order term disappears at $\Theta_{\text{MAP}}$. Then the posterior approximation $q(\Theta)$ can be identified as a Gaussian centered at $\Theta_{\text{MAP}}$, with a covariance corresponding to the local curvature: 
\begin{equation}
    q(\Theta) := \mathcal{N} (\Theta \mid \Theta_{\text{MAP}}, {({\nabla_\Theta^2 \mathcal{L} (\mathcal{D};\Theta)|}_{\Theta_{\text{MAP}}})}^{-1}).
\end{equation}
That is, the covariance is given by the inverse Hessian of the regularized training loss (which is interpreted as an unnormalized negative log posterior) at the trained weights $\Theta_{\text{MAP}}$.
 
A key practical advantage of this approach is that, since standard training of neural networks already identifies the local optimum $\Theta_{\text{MAP}}$, the only additional cost is
to compute the Hessian $\nabla_\Theta^2 \mathcal{L} (\mathcal{D};\Theta )$ at that point, once. 
This also means the approximation can be computed \emph{post-hoc}, for pre-trained networks, which implies that uncertainty quantification in the form of a Laplace approximation comes only at a very small computational overhead while also preserving the predictive power of the maximum a posteriori estimate.

As mentioned before, we can use the decomposition of the neural operator into a fixed feature map corresponding to the first $L-1$ layers and a last linear layer \citep{SnoekRSKSSPPA15}. This is particularly convenient in the case of the architecture considered by \citet{li2020neural}, since the last layer is indeed linear. Due to the linearity in the weights of the last layer, the distribution over the function outputs will also be Gaussian. Hence, for a Gaussian likelihood the predictive distribution in \Cref{eq:predictive} can be computed in closed form by using the approximate posterior $q(\Theta).$ Note that this predictive distribution is equivalent to the one of a GP regression problem \citep{khan19approximate}. This directly connects the GP approach for the shallow to the deep case, although we are now not approximating the posterior over the parameters of the Green function, but over the weights of the last layer.

\citet{kristiadi2020being, LaplaceRedux} recently showed that this approach achieves competitive performance on many common uncertainty quantification benchmarks compared to more recent alternatives -- despite the low computational overhead.
Last-layer Laplace approximations may seem like a simplistic approach to quantifying uncertainty of intricate architectures such as as neural operators. In \Cref{sec:experiments}, we show empirically that this is not the case.
%To adapt this approach on neural operator architectures (see again Figure \ref{fig:noparchitecture}), we can add a linear layer after the last layer $\psi_L$. prior With an isotropic Gaussian $p(W) = \mathcal{N}(0, \sigma^2 I)$ over the last-layer weights and a Gaussian likelihood we can compute the predictive distribution in closed form.  Our experiments show that this is indeed enough for good uncertainty estimates. 
%Alternatively, we can treat a subset of the neural operator weights probabilistically, apply the Laplace approximation to it,  and keep the rest of the weights to their MAP-estimates.
%We can for example choose to use a Laplace approximation on the weights $W_l$ for a chosen $l \in \{1, ...,L\}$. This approach requires approximation methods for the Hessian ().

% \begin{comment}
\section{RELATED WORK}\label{sec:related_work}
The interplay of (parametric) partial differential equation models (see \citet{cohen2015approximation} for a review) and deep learning has rapidly gained momentum in recent years.
Broadly speaking, there are two approaches: learning the solution of a given PDE on the one hand, and learning the parameter-to-solution operator of a family of parametric PDEs on the other hand.

Conventional numerical PDE solvers (e.g. \citep{ames2014numerical}) and physics-informed neural networks \citep{raissi2019physics,2018Sirign, zhu2019physics} fall into the first category.
In physics-informed neural networks, the PDE solution is modelled as a neural network.
The differential equation is then translated into an appropriate loss function, and an approximate PDE solution emerges from automatic differentiation and numerical optimisation.
While the physics-informed neural network formulation extends naturally to PDE inverse problems \citep{raissi2019physics,zhu2019physics}, it brings with it some practical issues like hyperparameter-sensitivity and complicated loss landscapes \citep{wang2021understanding,sun2020surrogate}.
Physics-informed neural networks also need to be retrained once the parametrisation of the PDE ($\lambda$ of $f$) changes. 
% {\color{red} \cite{li2021physics}  combines the physics-informed neural network approach and the operator learning approach. }

As described in \cref{sec:neural_operators}, neural operators do not face this issue because they learn the parameter-to-solution operator of a family of parametric PDEs (recall \cref{eq:neurop_generalized}).
They have been conceptualised by \citet{lu2021learning}, brought to the limelight by \citet{bhattacharya2020model,nelsen2021random,li2020neural,li2020multipole,li2020fourier,li2021markov,patel2021physics,duvall2021non,kovachki2021neural}, and further extended by \citet{guibas2021adaptive,gupta2021multiwavelet,rahman2022u,fanaskov2022spectral,rahman2022generative}.  They are equipped with uncertainty quantification in the present work.
The effect of uncertainty quantification will be investigated in the experiments below, and extending the proposed methodology will be an important direction for future development of neural operators because the solution operators are highly nonlinear, and its recovery strongly depends on the training data.
For sophisticated PDE models, training data is expensive to assemble because each training point relies on the numerical solution of a PDE.
In those low-data regimes, uncertainty quantification equips a practitioner with the information of whether the recovery can be trusted or not.

A principled approach to uncertainty quantification is generally provided by Bayesian deep learning. Besides the Laplace approximation, which has been discussed in \Cref{subsec:llla_neurop}, there are many more approximate Bayesian methods for inferring the neural networks' weights.
These include variational inference \citep{Graves2011PracticalVI, blundell2015weight, Khan2018FastAS, Zhang2018NoisyNG}, Markov Chain Monte Carlo \citep{neal1996, Welling2011BayesianLV, zhang2019cyclical}, and heuristic methods \citep{gal2016dropout, Maddox2019ASB}. 
Typically, they employ a Gaussian posterior approximation. 
One crucial advantage of the Laplace approximation over many of these methods is that it can be applied \textit{post-hoc}, i.e. it is not only cheap but also preserves the estimate returned by the preceding non-Bayesian computation, whereas other methods require retraining the network, which is expensive and might lead to worse predictive performance, since the optimization process is changed and the standard deep learning techniques and hyperparameter settings might not work anymore, which makes more tuning necessary; this is an additional cost factor besides the training itself.

% Recently, different machine-learning-based methods for solving PDEs have been developed. The main approaches are either to  approximate the solution function of the PDE (\cite{raissi2019physics,2018Sirign, zhu2019physics} )
% or learning directly the solution operator over a family of PDEs.
% Other deep learning methods directly learn the Green function ()
% \end{comment}

\section{Experiments}
\label{sec:experiments}
In this section we exploit the theoretical analysis developed in \Cref{method} to  construct Bayesian neural operators delivering uncertainty estimates. 
We use the analytic GP framework of \cref{subsec:gp_regression} to build a non-parametric Bayesian neural operator in the "shallow" case, then extend our method to the deep case. We reproduce the experiments on  neural operators as carried out by \cite{li2020neural} to  show that we can effectively detect wrong predictions. 

\subsection{Uncertainty Quantification in the Shallow Case with GP regression}\label{subsec:gp_experiments}

Consider the boundary value problem in \cref{eq:test_problem} for a fixed $\lambda_0 \in \Rbb$. As discussed in \cref{subsec:gp_regression}, since the PDE is linear and admits the Green's function $G \colon \Rbb^2 \rightarrow  \Rbb$ in \cref{eq:greens_function}, inferring the solution operator $\mathcal{G} \colon f \mapsto u$ is  equivalent to learning the function $G$ given integral observations  ${\{f_i,  u_i = \int_D G(\cdot, y) f_i(y) \diff y \}}_{i=1}^N$. Note that every observation point is a function, numerically observed on a grid $\Xbb = \{x_1, \dots x_K \}$. As training points $\{f_i \}_{i=1}^N$ (right hand functions of the PDE in \cref{eq:test_problem})  we use the first $N$ Legendre polynomials shifted on the interval $[0, 1]$ and observed on an evenly spaced grid $\Xbb = \{x_1=0, \dots x_9=1 \}$. 
We assume a Gaussian prior $G \sim \mathcal{GP}(\mu, k) $ with a zero mean function $\mu$ and a kernel function $k \colon \mathbb{R}^2 \times \mathbb{R}^2 \rightarrow \mathbb{R} $ that factorizes into the product $k((x_0, x_1), (y_0, y_1)) = k_1(x_0, y_0) k_2(x_1, y_1)$ where $k_1$ and $k_2$ are  Mat\'ern kernels with parameter $\nu = 2.5$. To compute the integral operator $\mathcal{A}$ in \cref{eq:gp_post} we use numerical integration. \\

\begin{figure*}[ht] 
      \begin{center} 
            %\framebox[4.0in]{$\;$}
            \includegraphics[width=\textwidth]{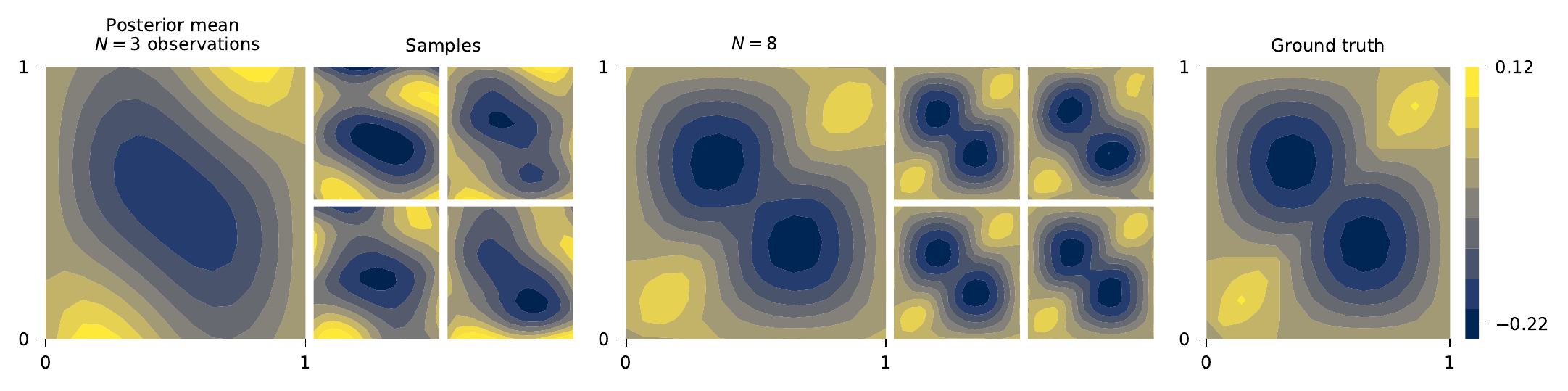}
      \end{center}
      \caption{Posterior distribution on  $G_{\lambda_0}$ for $\lambda_0 = 4.5$ (and ground truth) after $N= 3,8$ observations $\{f_i \}_{i=1}^N$ with $f_i $ shifted Legendre polynomials. The samples show the approximation's variance, which decreases when $N$ increases. }\label{fig:gp_post}
\end{figure*}

The posterior distribution over $G$ inferred with \cref{eq:gp_post} is shown in \cref{fig:gp_post}.
The figure shows the posterior distribution for $G$ after $N=3$ and $N=8$ function observations.
% Since  $G \colon \Rbb^2 \rightarrow  \Rbb$,
Samples from the posterior are used to visualize the posterior variance;   the very diverse samples after  $N=3$ observations derive from a high variance, reflecting the fact that the approximation is  imprecise. Instead, for $N=8$ the distribution exhibits low variance and an accurate estimate. Since by learning $G$ we learned the inverse of the differential operator in \cref{eq:test_problem},  we can use the posterior distribution over $G$ to get an approximation, as well as an error estimate, of the solution for a new PDE with right hand side function $f^\ast$.  

\subsection{Uncertainty Quantification in the Deep Case }
 The examples considered in this section aim to emphasize the importance of uncertainty quantification
for the complex architectures that are neural operators. 
We show that Bayesian neural operators are able not only to precisely detect areas
of imprecise solution estimates, but in low
sampling regimes they help preventing
significant mistakes in the prediction.
The fact that such mistakes can occur already in linear PDEs, which we consider in this section, further motivates the need for uncertainty estimates in the nonlinear case.

To recreate the results in \citet{li2020neural} we use their original code.\footnote{\href{https://github.com/zongyi-li/graph-pde/tree/08ab9d8f8e9d1a6d95017a6f803a53d8a603e66e/graph-neural-operator}{\color{blue}https://github.com/zongyi-li/graph-pde/graph-neural-operator}}
%We emphasize that the neural operator implementation of \cref{neurop_iteration} considered in this section
%is exactely the one from their work. 
As discussed in \cref{subsec:llla_neurop}, our Bayesian framework computes Gaussian approximations of the posterior $p(\Theta \mid D)$ through Laplace approximations.
For an efficient implementation of the last-layer Laplace approximation, we use the software library introduced by \cite{LaplaceRedux}. 
%As described in \Cref{subsec:llla_neurop}, 
We use a last-layer Laplace approximation with a full generalized Gauss-Newton approximation \citep{Schraudolph02} of the Hessian. There are two scalar hyperparameters, the prior precision and the observation noise. Both are tuned \textit{post hoc} via optimizing the log marginal likelihood \citep{immer2021scalable, LaplaceRedux}.

In their work, \cite{li2020neural}
considered the second order elliptic PDE 
\begin{equation}
\begin{split}\label{darcyflow}
- \nabla \cdot (\lambda(x) \nabla u(x)) & = 
f(x), \quad x \in D \\
u(x) & = 0 \ \  \qquad x \in \partial D
\end{split}
\end{equation}
where $D = [0,1]^2$ is the unit square and $f \equiv 1$. The PDE in \cref{darcyflow} represents the steady state of a two dimensional Darcy flow and arises in several physical applications. The nonlinear solution operator 
\begin{equation}
 \mathcal{F} \colon \Lambda \rightarrow U, \quad \lambda \mapsto u
 \end{equation}
 is approximated with a type of neural operator architecture based on graph neural network structures (\cite{kipf2016semi}). In particular, for the computation of
 the integral in \cref{neurop_iteration}, the  domain $D$ is discretised into a graph-structured data on which the message passing algorithm of \cite{pmlr-v70-gilmer17a} is applied. 
 
%\textit{Here: detail of the Bayesian framework. last layer laplace, tuning of hyperparameters?}
 
In Section \ref{small_data_grid} we examine the case
where only few data are available. In Section 
\ref{big_data_grid} a high data regime is considered.

\begin{figure}[t]
    \centering
    \begin{subfigure}[h]{0.5\textwidth}
        \centering
        \includegraphics[width=\textwidth]{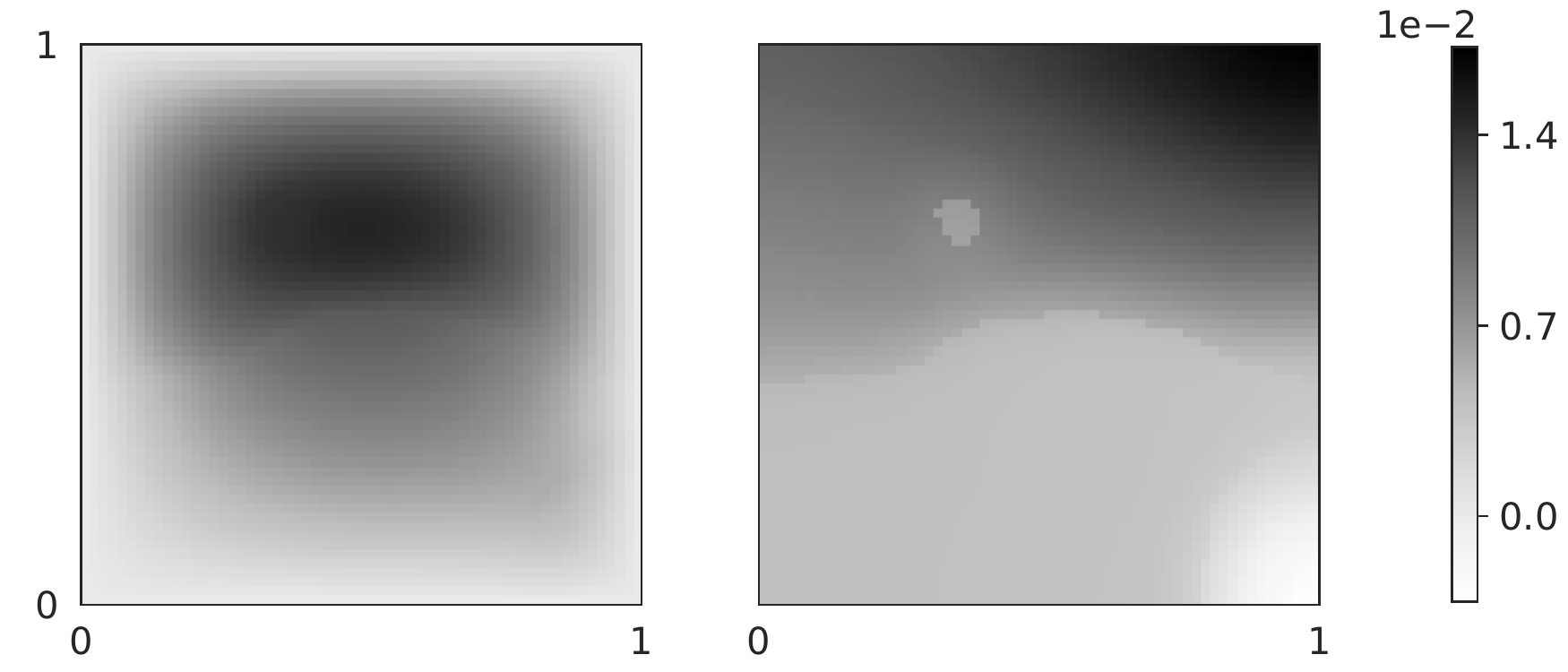}
        \caption{The ground truth (\textbf{left}) and the approximation (\textbf{right}).}
        \label{subfig:lowdata_truth_approx}
     \end{subfigure}
     
    % \vspace{2em}
     
     \begin{subfigure}[h]{0.5\textwidth}
        \centering
        \includegraphics[width=\textwidth]{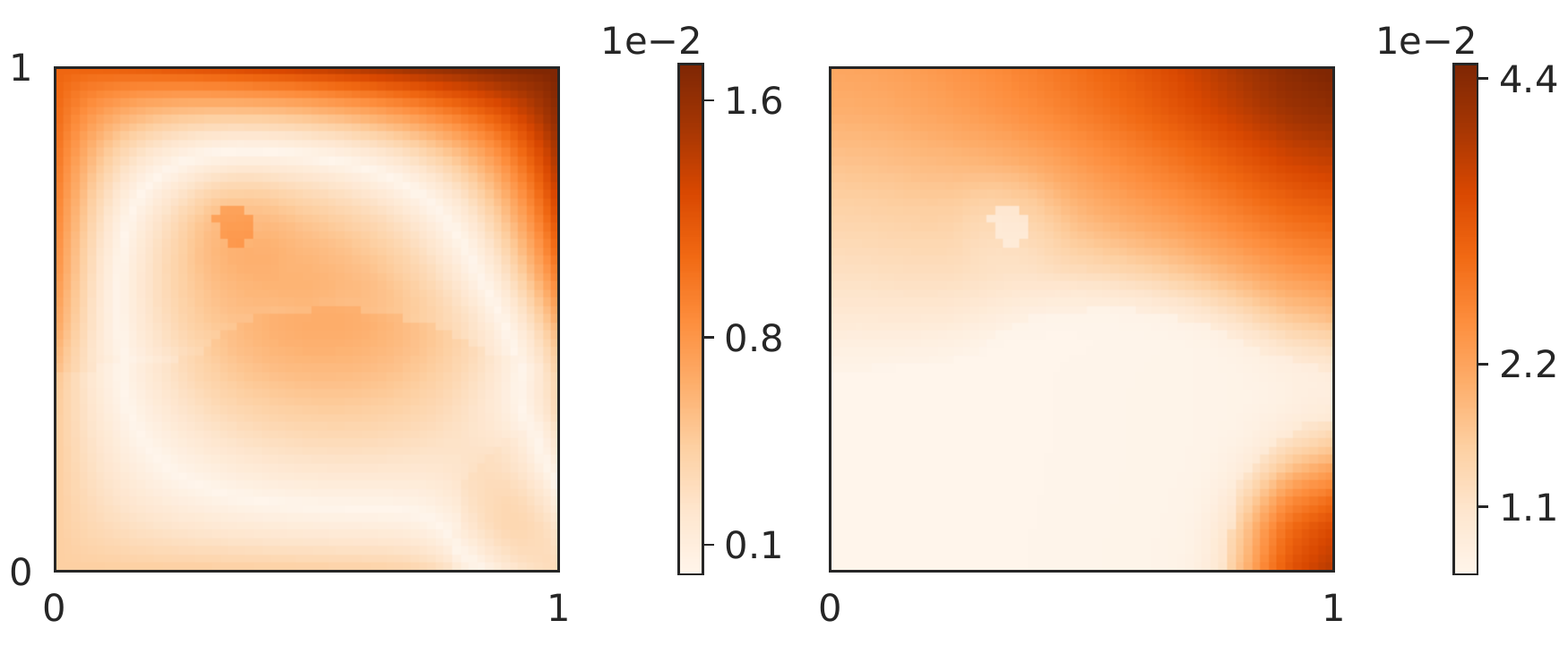}
        \caption{The error (\textbf{left}) and the standard deviation (\textbf{right}).}
        \label{subfig:lowdata_err_std}
     \end{subfigure}
    \caption{The Bayesian neural operator on the 2d Darcy flow problem in the low-data regime. The approximation is relatively bad and the predictive standard deviation highlights the areas of high error.}
    \label{fig:lowsamplesregime}
\end{figure}

\subsection{Low-data Regime}\label{small_data_grid}
At first, we consider
the case where only few observation points on the unit square $D =[0,1]^2$ 
are available. 
Sparse observations are a common scenario for multi-scale dynamics described by PDEs, where 
data is usually not easily affordable.  Due to the limited amount of data, approximated solutions can be very inaccurate.
It is therefore fundamental to
associate uncertainty to the outcome of the 
prediction.

%\textit{Here: Details of the experiment: how many grid points per input function, subsampling points}
In particular, since the problem is relatively simple, we consider an extreme setting where we train on only two training functions and subsample only two points from a $16 \times 16$ grid for each.
\Cref{fig:lowsamplesregime} shows on a $61 \times 61$ grid that in this setting the neural operator  fails to predict the solution well. As a consequence, the Bayesian neural operator yields low confidence (high predictive standard deviation) in the prediction, particularly in the areas of higher error. For readability, the plots use different color scales. This is due to the slight underconfidence of the Laplace approximation (in the scalar global parameter, not the local structure).
Having measures such as the predictive standard deviation to determine whether the prediction should be trusted is of big practical benefit for many applications.

\subsection{High-data Regime}\label{big_data_grid}
The previous section considered a heavily under-sampled problem. In this simple uni-variate toy problem, this meant using a very small number of training data. Nevertheless, the under-sampled regime is arguably the norm in practical problems with a more realistic, higher number of dimensions, where one can not hope to tile the domain with pre-computed PDE solutions. In this section, for completeness, we consider the other, over-sampled end of the spectrum and find that good and structured uncertainty quantification is nevertheless useful here.

\begin{figure}[t]
    \centering
    \begin{subfigure}[h]{0.5\textwidth}
        \centering
       
        \includegraphics[width=\textwidth]{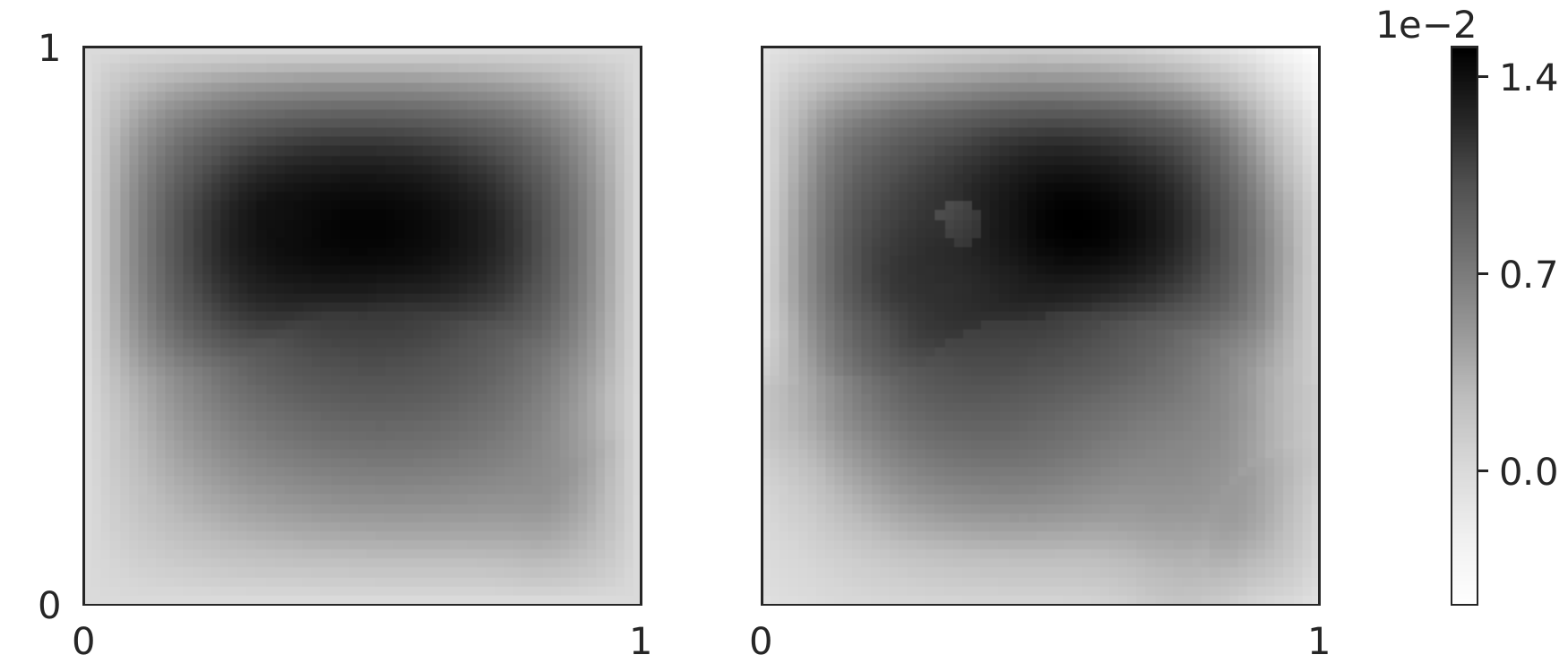}
        \caption{ The ground truth (\textbf{left}) and the approximation (\textbf{right}).}
        \label{subfig:truth_approx}
     \end{subfigure}
     
   %  \vspace{2em}
     
  \begin{subfigure}[h]{0.5\textwidth}
        \centering
        \includegraphics[width=\textwidth]{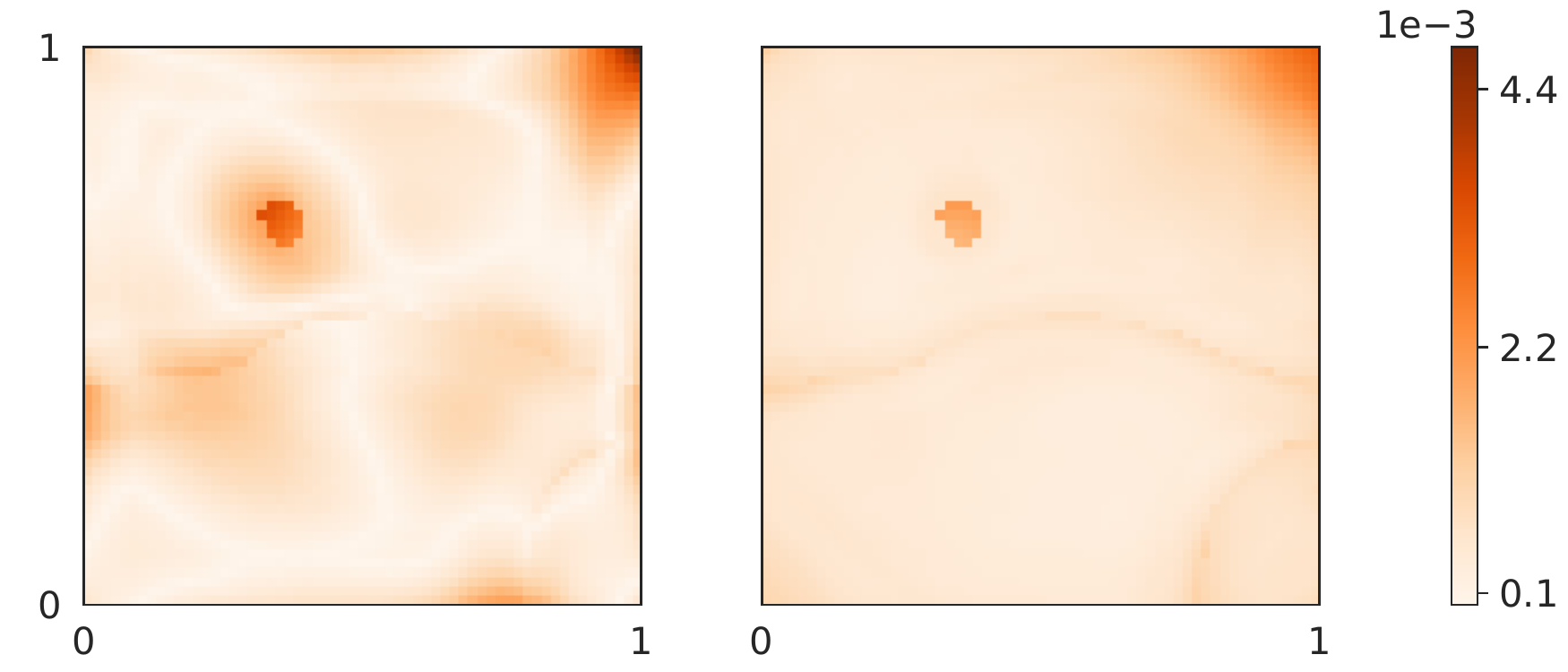}
        \caption{The error (\textbf{left}) and the standard deviation (\textbf{right}).}
        \label{subfig:err_std}
     \end{subfigure}
    \caption{The Bayesian neural operator on the 2d Darcy flow problem in the high-data regime. The approximation is close to the ground truth. The regions of relatively high error, as well as their magnitude, are captured by the predictive standard deviation.}
    \label{fig:highsamplesregime}
\end{figure}

\Cref{fig:highsamplesregime} shows results on a dense $61 \times 61$ grid, analogous to the previous one, trained on 100 densely evaluated $16 \times 16$ grid solutions. Note, that the model generalizes well from the smaller $16 \times 16$ grid used during training to the larger $61 \times 61$ grid for testing, as previously shown by \cite{li2020neural}.
Although the prediction error is generally of good quality (i.e.~relative prediction errors are mostly below 10\%), the trained network exhibits an artifact in one, sharply delineated region of the training domain. This is not a bug, but a common problem with the ReLU features in this architecture, which create piecewise linear predictive regions \citep{hein2019relu}. 

As the figure shows, the Laplace approximation is in fact able to identify and delineate this region well, and produce an effective, well-calibrated warning about its presence. It is important to note that this kind of functionality is only possible with the \emph{structured} uncertainty produced by a Bayesian technique like the Laplace approximation -- i.e.~by an approximate posterior measure, rather than a global worst-case error bound.

\begin{comment}
\begin{figure}[h]
            %\framebox[4.0in]{$\;$}
            \includegraphics[width=.5\textwidth]{full_grid_s16_n100.pdf}
            \caption{\textbf{First row:} ground truth (\textit{left}) and approximated solution (\textit{right}). \textbf{Second row:} error (\textit{left}) and standard deviation (\textit{right}).\emilia{put plot where error is also on the bottom right corner}}\label{fig:highsamplesregime}
\end{figure}
\end{comment}

\section{CONCLUSIONS}
 We provided a theoretical Bayesian framework for neural operators. While these recently 
 introduced architectures have shown to
 have a competitive performance with other numerical
 methods and to beat the
 state-of-the-art neural networks approaches on 
 large grids, they did not previously come with explicit uncertainty quantification. 
 We developed an explicit analytic Bayesian treatment for the linear base-case, and illustrated how we can learn (the distribution over) solution operators through non-parametric GP regression. We provided an effective and efficient approximate Bayesian treatment for the full, deep case through the use of Laplace approximations. 
 In experiments, our approach is able to quantify predictive uncertainty both in the sparsely and densely sampled regime. In the former, it produces structured uncertainty across the predictive domain. In the latter, it is able to precisely detect and delineate regions where the predictive estimate fails to approximate the true solution well. The code used to produce the results herein will be released with the final version of this paper. 

 If deep learning approaches to the simulation of dynamical systems are to fulfill their potential and be applied to serious, large-scale partial differential equations (including safety-critical and scientific applications), then uncertainty quantification as presented here has a crucial role to play in the prevention of accidental and potentially dangerous prediction errors.
\newpage
\section*{Acknowledgments}
E.M, N.K., R.E. and P.H. gratefully acknowledge financial support by the European Research Council through ERC StG Action 757275 / PANAMA; the DFG Cluster of Excellence “Machine Learning - New Perspectives for Science”, EXC 2064/1, project number 390727645; the German Federal Ministry of Education and Research (BMBF) through the Tübingen AI Center (FKZ: 01IS18039A); and funds from the Ministry of Science, Research and Arts of the State of Baden-Württemberg. E.M, N.K., R.E. and P.H. gratefully acknowledge financial support by the German Federal Ministry of Education and Research (BMBF) through Project ADIMEM (FKZ 01IS18052B).  E.M. is grateful to the International Max Planck Research School for Intelligent Systems (IMPRS-IS) for support.
L.R. acknowledges the financial support of the European Research Council (grant SLING 819789), the AFOSR project FA9550-18-1-7009 (European Office of Aerospace Research and Development), the EU H2020-MSCA-RISE project NoMADS - DLV-777826, and the Center for Brains, Minds and Machines (CBMM), funded by NSF STC award CCF-1231216.

\bibliography{uai2022-template}
\end{document}